\documentclass{article}
\usepackage{bm}
\usepackage{bbm}
\usepackage{tabularx,booktabs}
\usepackage{amsfonts}
\usepackage{spconf,amsmath,graphicx}

\title{Instance segmentation with the number of clusters incorporated in embedding learning}
\name{Jianfeng Cao, Hong Yan}
\address{Department of Electrical Engineering, City University of Hong Kong, Hong Kong, China}
\begin{document}
%
\maketitle
%

\begin{abstract}
Semantic and instance segmentation algorithms are two general yet distinct image segmentation solutions powered by Convolution Neural Network. While semantic segmentation benefits extensively from the end-to-end training strategy, instance segmentation is frequently framed as a multi-stage task, supported by learning-based discrimination and post-process clustering. Independent optimizations on substages instigate the accumulation of segmentation errors. In this work, we propose to embed prior clustering information into an embedding learning framework FCRNet, stimulating the one-stage instance segmentation. FCRNet relieves the complexity of post process by incorporating the number of clustering groups into the embedding space. The superior performance of FCRNet is verified and compared with other methods on the nucleus dataset BBBC006.  
\end{abstract}
%

\section{Introduction}
\label{sec:intro}
Automatic segmentation tasks can be generally divided into two groups, semantic and instance segmentations. While semantic segmentation discriminates front objects from the background, instance segmentation dives deeper to delineate each object in the front mask with different labels. Recently, substantial efforts have been made to semantic segmentation in various areas, such as the automatic annotation on cardiac and brain \cite{tu2008brain}. For these binary or multi-class segmentation tasks, the learning target can be easily formulated as multi-channel matrix, in which each channel corresponds to one specific class. However, the instance segmentation can be hardly achieved by encoding different objects in a similar way. First of all, the number of objects varies with different samples. This means that if each object is represented by one channel, the number of output channels should change adaptively according to different inputs, leading to dynamic weights in the trained model. Even though the output channels can be theoretically set as the maximum number of objects to accommodate all cases, the computational resource becomes practically demanding. Secondly, given two similar objects in the same class, the model is supposed to assign them to different channels. Sometimes this requirement is ill-defined when even humans can hardly define the specific proxy labels for two similar objects.

Previous studies have reported important insights to mitigate the difficulties in instance segmentation. One category is to treat the instance segmentation as multi-stage or multi-task process. Watershed segmentation is a widespread algorithm to split the region into different partition. Because over-segmentation error exists commonly in maximum-seeded transformation, deep learning is utilized to derive the seeds and binary mask or even the distance map individually  \cite{koohbanani2019nuclear, chen2016dcan, vuola2019mask}. After decomposing the instance segmentation into subtasks, the improvements in classification or regression can be incorporated into the pipeline. Unfortunately, errors in suboptimal systems will accumulate in downstream post-process when the pipeline is decoupled and cannot be optimized integrally. 

The possibility of instance segmentation also emerges from the embedding learning. Embedding learning evolves from the deep matrix learning, which is commonly applied to person re-identification and image retrieval. With all inputs, from either the same class or different classes, represented as a feature vector by the neural network, the main objective of deep matrix learning is to maximize within-class similarity and minimize between-class similarity. Similarly, embedding learning encodes each pixel instead of the whole input into a higher feature space, where pixels tend to be self-organized under the supervision of similarity based loss functions \cite{payer2018instance, chen2019instance}. To mitigate the difficulty in optimization, objects can be designated with the combination of fixed feature maps \cite{kulikov2020instance}. Subsequently, the model is trained to imitate this process and bridge the direct transformation from the raw image to the coefficient map. Following the embedding learning, clustering plays an essential role in furnishing the embedding map. Although several studies present \textit{ad hoc} methods to elevate the general performance on clustering embedding vectors into different instances, such as jointly representation learning and clustering \cite{fard2020deep, yang2017towards}, the model is mostly trained through alternating stochastic optimization. In order to optimize the model globally, there is an increasing demand in instance segmentation model suitable for end-to-end training. 

In light of the embedding learning, diverse attempts have been made to enhance the descriptiveness of the feature vector, meanwhile facilitating the clustering performance. For each pixel, the output vector is now composed of multiple quantities with morphological meanings, such as centre location and object size, as opposed to naive high-dimensional point used for clustering. For example, CenterMask assembles the final segmentation by using size branch to tune the binary shape mask, along with a heatmap branch to filter object centres \cite{wang2020centermask}. SOLO framework provides an impressive solution for one-stage instance segmentation \cite{wang2019solo, wang2020solov2}. Its category branch outputs which pixel belongs to which class, and the mask branch dynamically outputs the corresponding object mask. In these methods, opportunistic policies are well defined to curtail the post-processing stage, such as clustering, whose optimum status is imperceptible to the model optimizer. However, predicting the centre, especially in densely packed objects (e.g., cells), is a non-trivial task.

This paper discusses a simple yet effective one-stage instance segmentation framework FCRNet by combining the four color theorem and embedding learning strategy. As the four color theorem states that four colors are enough for separating adjacent regions, FCRNet projects all pixels to only four groups in the feature space. The constraint on the number of cluster is performed by the argmax activation, a parameterized version of softmax function. The model FCRNet refrains from complex downstream processes (e.g., mean-shift or $k$-means clustering) because the number of embedding groups during the training stage is explicitly specified. The embedding map can be easily converted to instance segmentation according the regional connectivity. To substantiate its superiority, FCRNet is tested and compared with other methods on the public nucleus dataset BBBC006.

\section{Method}
Compared to semantic segmentation, the difficulty of instance segmentation can be profiled from two aspects, the flexible number of objects and implicit intra-class difference. In this work, given the input composed of multiple objects, on the one hand embedding learning outputs the embedding maps, where pixels belonging to different objects can be easily differentiated; on the other, a hardmax activation is approximated with parameterized softmax in order to regulate the groups in the end-to-end training. 

\begin{figure}[htb]
    \begin{minipage}[b]{0.45\linewidth}
      \centering
      \centerline{\includegraphics[width=4.0cm]{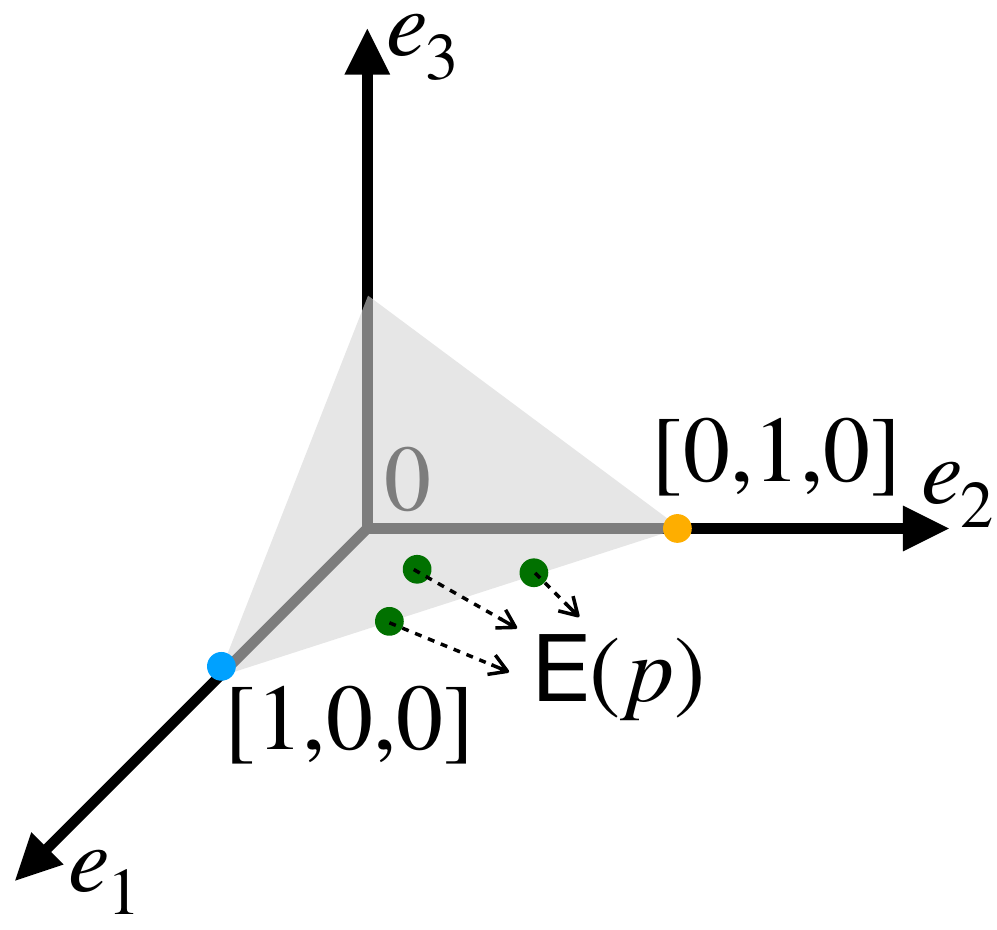}}
      \centerline{(a)}\medskip
    \end{minipage}
    \hspace{0.5cm}
    \begin{minipage}[b]{0.45\linewidth}
      \centering
      \centerline{\includegraphics[width=4.0cm]{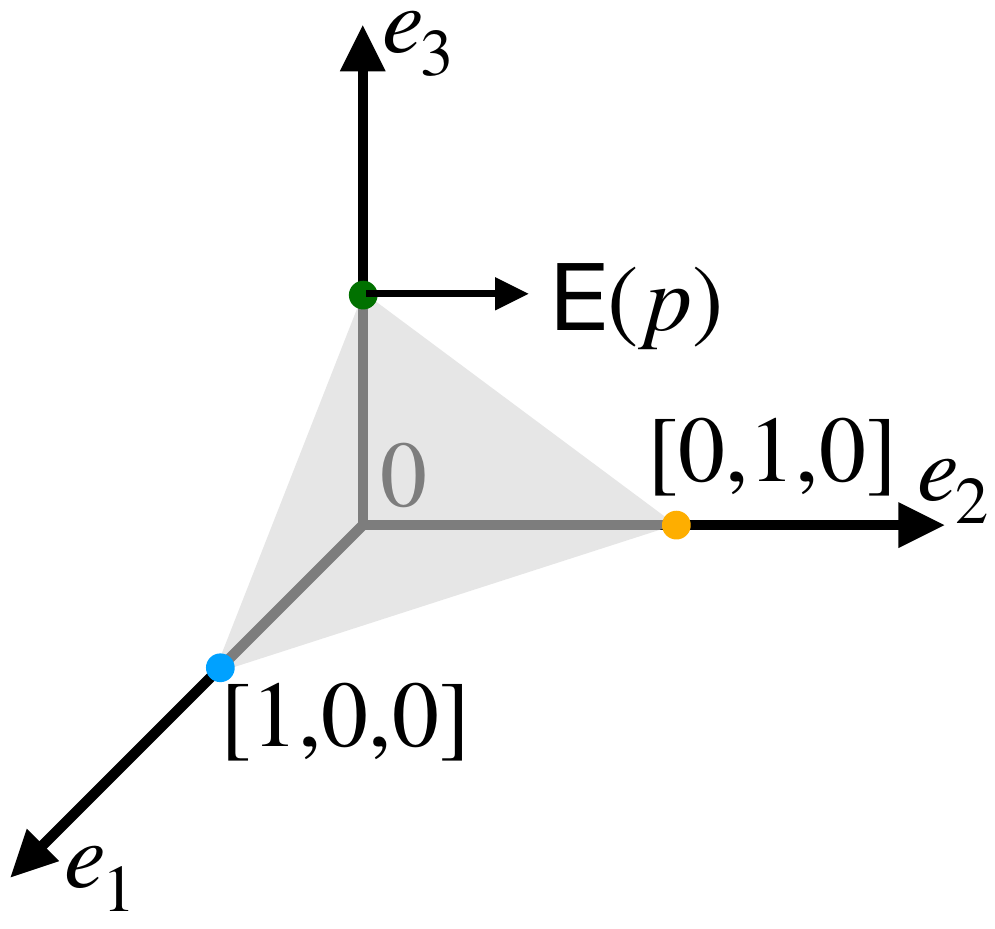}}
      \centerline{(b)}\medskip
    \end{minipage}

    \caption{The influence of the constraint on the number of cluster centres. (a) For a new class (green) to be added, it can distribute at multiple locations inside the simplex as long as it keeps away from other two classes (aqua and yellow). (b) Under the constraint, the new class is limited around the vertex. $K=3$ in this example.}
    \label{fig:constrain}
\end{figure}

\subsection{Embedding Learning}
Supposing the input $\bm{\mathrm{I}}: \Omega \subset \mathbb{N}^2 \rightarrow \mathbb{R}$ is composed of $M$ objects $\bm{\mathrm{I}}=\{\bm{\mathrm{O}}_i | i\in {1, 2, ..., M}\}$, the embedding network $\phi_{\theta}: \bm{\mathrm{I}} \rightarrow \bm{\mathrm{E}}$ maps each pixel to a $K-$dimensional vector in space $\bm{\mathrm{E}}:\Omega \rightarrow \mathbb{R}^K$, where $\theta$ denotes the parameters of the network. Pair-wise similarity $g(\cdot, \cdot)$ can be used to measure the similarity of two pixels in $\bm{\mathrm{E}}$ based on distance functions, such as cosine and Euclidean distances. Embedding learning aims at finding the optimal model $\theta^*$ by solving:
\begin{equation}
\label{eq1}
\left\{
    \begin{aligned}
        \theta^* & = \arg\mathop{\max}\limits_{\theta}\text{ } l_{\text{intra}}(\bm{\mathrm{E}};\theta) - l_{\text{inter}}(\bm{\mathrm{E}};\theta) \\
        l_{\text{inter}} & = \sum\limits_{i\in \{1, 2, ..., M\}}\frac{1}{|N(i)|}\sum\limits_{j\in N(i)} g(\bm{\mu}(i), \bm{\mu}(j))\\
        l_{\text{intra}} & = \sum\limits_{i\in \{1, 2, ..., M\}} \quad \sum\limits_{p\neq q|p,q\in \bm{\mathrm{O}}_i} g(\bm{\mathrm{E}}(p), \bm{\mathrm{E}}(q))
    \end{aligned}
\right.
\end{equation}
where $\bm{\mu}(\cdot)=\frac{1}{|\bm{\mathrm{O}}_{\cdot}|}\sum_{p\in \bm{\mathrm{O}}_{\cdot}}\bm{\mathrm{E}}(p)$ is the mean feature of one object. $N(\cdot)$ represents the neighbors of a object. The solution $\theta^*$ stimulates the system to reach a status where between-class similarity $l_{\text{inter}}$ is minimized and within-class similarity $l_{\text{intra}}$ is maximized. Consequently, pixels in the same object will be grouped together. At the same time, pixels from different objects will not be included. 

\subsection{Learning with Constrained Number of Clusters}
Clustering analysis is widely applied to the embedding feature $\bm{\mathrm{E}}$. Based on the availability of the number of clusters or objects, either centre-based or density-based clustering is applicable. For example, $k$-means clustering is preferable when an additional head branch is attached to predict the number of instances. Alternatively, mean-shift is well suited for processing self-organized points. In these methods, the clustering is independent from the training on the network model, which means that $\phi_{\theta^*}$ can guarantee only the fulfillment of policies on intra-or-inter class similarity but not the quality of clustering results. The instance segmentation, however, holds the assumption that the embedding space matches well with the clustering method, which is deceptive when these two stages are not trained together. 

Here, we attempt to specify the number of embedding groups during the training. In Eq.(\ref{eq1}), the embedding vector $\bm{\mathrm{E}}(p)$ of pixel $p$ locates flexibly inside a simplex, scattering the potential cluster centres. Taking $K=3$ in Fig.\ref{fig:constrain}  as an example, $\bm{\mathrm{E}}(p)$ could be any point between clusters $[1,0,0]$ and $[0,1,0]$ (e.g., $[0.5,0,0]$) as long as it fulfills the similarity requirements. In other words, every time there comes a new object, an extra cluster centre would possibly emerge. It severely increases the clustering uncertainty and complicates the subsequent processes where over-or-under segmentation turns into the prime consideration. 

To overcome this problem, the embedding $\bm{\mathrm{E}}(p)=[e_1, e_2, ..., e_K]$ in this work is forced to be chosen from a finite set through a argmax activation
\begin{equation}
    \label{eq:argmax}
    h(\bm{\mathrm{E}}(p))_k:= \left\{ 
    \begin{array}{cl}
         1& \text{if }k=\arg \mathop{\max}\limits_j\text{ } e_j\\
         0& \text{otherwise}
    \end{array}
    \right.
\end{equation}
Theoretically, an ideal argmax activation is non-differentiable and cannot support the training. In order to promote the propagation of gradient, a parameterized softmax is designed as the output layer. Conventional softmax activation $f(\cdot )$ outputs the probability of each class defined as
\begin{equation}
\label{softmax}
    f(\bm{\mathrm{E}}(p))_k := \frac{\exp e_k}{\sum_{j=1}^K \exp e_j} 
\end{equation}

Similarly, the argmax $h(.)$ in Eq.(\ref{eq:argmax}) can be approximated with
\begin{equation}
\label{eq:argmaxapprox}
    h(\bm{\mathrm{E}}(p))_k \approx \frac{e_k^{\alpha}}{\sum_{j=1}^K e_j^{\alpha}}, \quad \alpha \in \mathbb{N}^+
\end{equation}
where $\alpha$ controls the concentration around the largest value. When $\alpha$ becomes larger, Eq.(\ref{eq:argmaxapprox}) approximates the argmax function better and still retains the gradient propagation. 

\begin{figure}[htb]
    \centering
    \centerline{\includegraphics[width=1.0\linewidth]{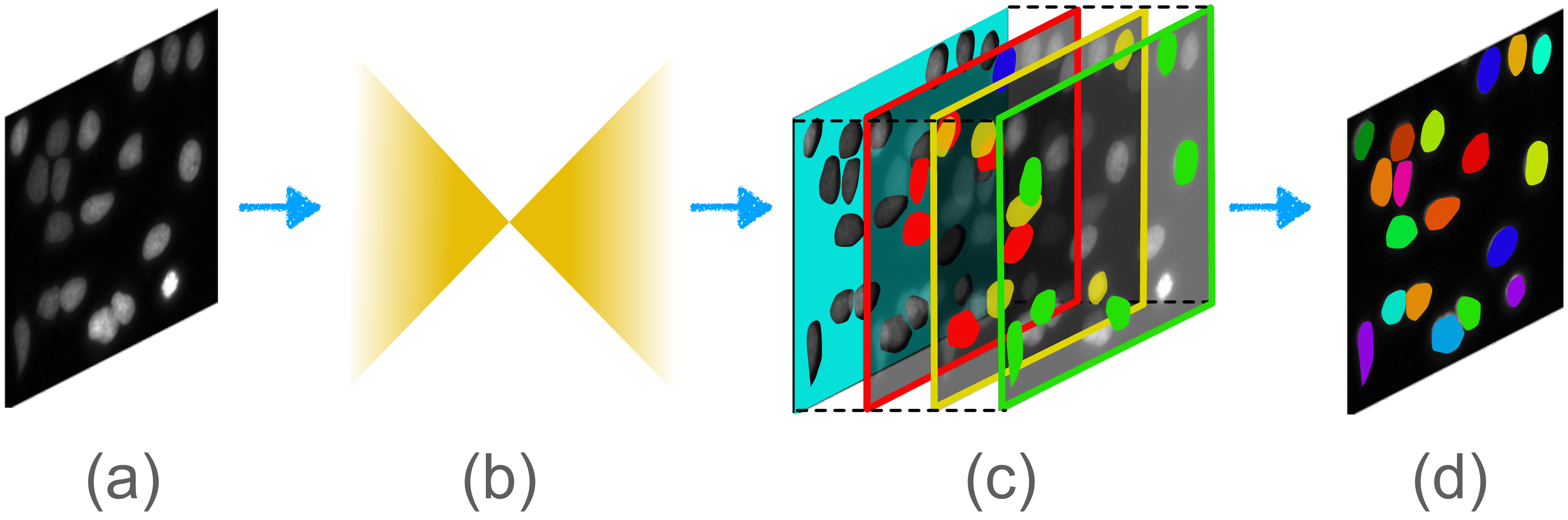}}
    \caption{A diagrammatic explanation of embedding learning constrained by argmax function. (a) Raw nucleus image is processed by (b) FCRNet equipped with argmax activation at the final layer. While the (c) output is composed of four channels, regions can be labelled according to the connectivity in each channel separately. All the objects contribute to the  (d) final instance segmentation.}
    \label{fig:four}
\end{figure}

Next, we need to decide the number of groups imposed on the embedding learning process. Given a separation of a plane, the four color theorem states that four colors are sufficient to color any adjacent regions with different colors. Therefore, the dimension is correspondingly set as $K=4$ so that Eq.(\ref{eq:argmaxapprox}) generates only four embedding vectors, i.e. $[1, 0, 0, 0]$, $[0, 1, 0, 0]$, $[0, 0, 1, 0]$ and $[0, 0, 0, 1]$, possibly with small approximation error (Fig.\ref{fig:four}). 

\begin{table*}
\centering
\caption{PERFORMANCE EVALUATION OF DIFFERENT METHODS}
\label{table:result}
    \begin{tabularx}{0.7\linewidth}{cccccc}
        \addlinespace
        \toprule
        Method     & Dice2      & AJI       & F1-score      & PQ    & \#Parameters \\ 
        \midrule
        DCAN \cite{chen2016dcan}   & 0.6341        & 0.4748         & 0.9059           & 0.4887     & 28.0M \\ 
        Discriminative Loss \cite{de2017semantic}  & \textbf{0.7594}        & 0.6763        & 0.8582           & 0.7458    &    \textbf{2.6M}\\ 
        Harmonic Embedding \cite{kulikov2020instance}    & 0.7305       & 0.6768          & \textbf{0.9574}      & 0.6827   &     43.0M \\ 
        FCRNet\_3 & 0.7035 & 0.7171 & 0.9232 & 0.7350 & 2.7M \\
        FCRNet\_10 & 0.6733 & 0.6271 & 0.7970 & 0.6596 & 2.7M \\
        FCRNet (ours)       & 0.7136        & \textbf{0.7513}          & 0.9294   & \textbf{0.7753}  & 2.7M \\ 
        \bottomrule
    \end{tabularx}
\end{table*}

\begin{figure}[htb]
\centering
    \centerline{\includegraphics[width=1.0\linewidth]{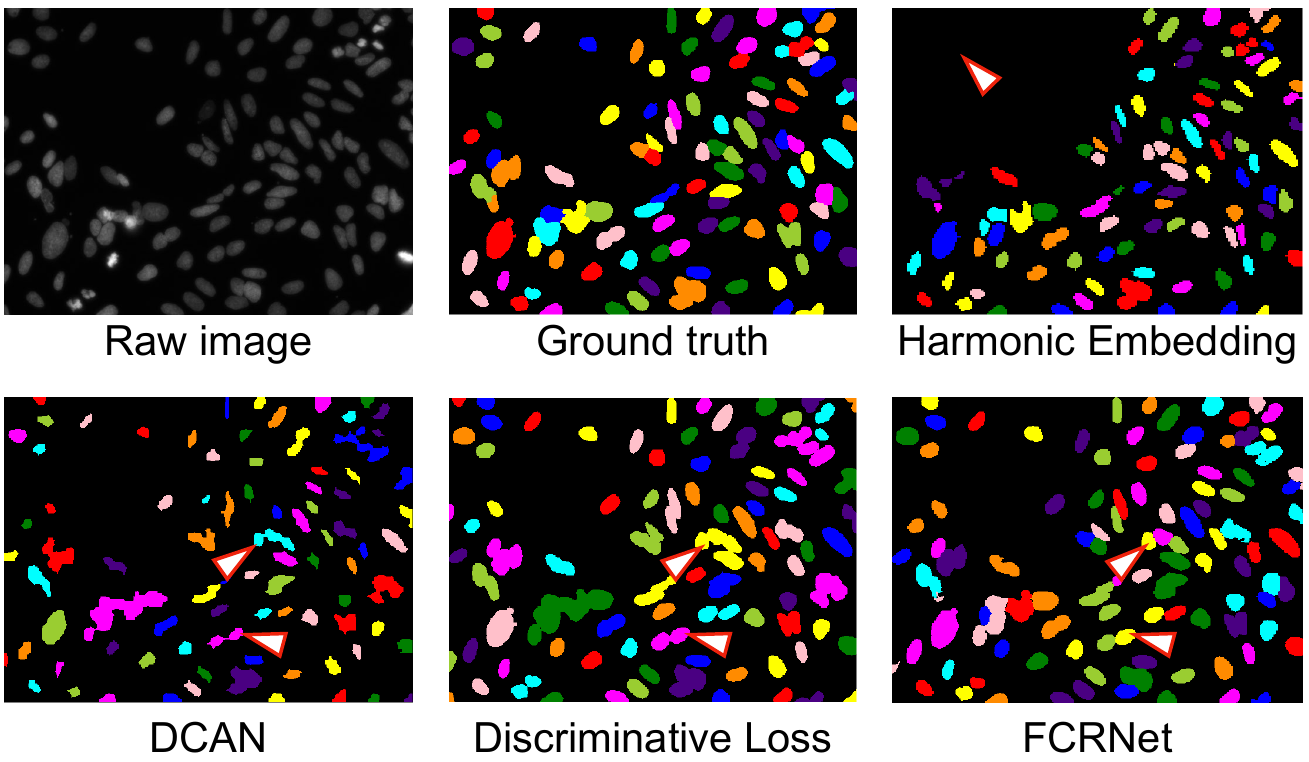}}
    \caption{Segmentation examples of compared methods. Some noticeable differences are marked with white triangles.}
    \label{fig:compare}
\end{figure}

\subsection{Postprocessing}
The constraints on the embedding space allow a fast post processing. Based on the output of $h(\cdot)$, all pixels can be easily grouped into four classes, within which neighboring objects are assigned to different channels. Objects can be further recognized in each class according to the regional connectivity. Finally, the instance segmentation is composed of all objects derived from four classes.

\section{Experiments}
\textbf{Dataset}. We tested the method on BBBC006 dataset \cite{ljosa2012annotated}, which includes 768 human U2OS cell sample images of $696\times 520$ pixels at different focal planes. Only images at $16^{\text{th}}$ plane were used in this study. The dataset was divided into 604 training and 164 evaluation images. While all images were normalized before input to the network, the images for training were further resized to $512\times 512$.

\textbf{Network structure}. FCRNet adopted the original U-Net architecture \cite{ronneberger2015u} except the final output layer, which was constructed based on the argmax function in Eq.(\ref{eq:argmaxapprox}). FCRNet used 16 convolution filters in the first layer and doubled this number in the following four submodules. The information in each submodule flowed through a Conv2D-Conv2D-ReLU way. For the final layer, the activation layer was replaced by Eq.(\ref{eq:argmaxapprox}) directly. Besides, we also tested two variants of FCRNet by changing the number of output channels to 3 and 10, namely FCRNet\_3 and FCRNet\_10 respectively. Maxpooling and linear upsampling were used in the encoding and decoding stages. In order to train the FCRNet, Adam optimizer was employed with a batch size of 4 and a initial learning rate of $1\times 10^{-4}$. The embedding similarity was set based on the cosine distance. The model was trained for 600 epochs and the learning rate was decreased by 0.9 exponentially for every 80 epochs. The approximation of  Eq.(\ref{eq:argmaxapprox}) was phased by increasing $\alpha$ gradually at different training stages. In our case, $\alpha$  was changed to $[2.0, 2.0, 4.0, 6.0, 8.0]$ after every 80 epochs.

To demonstrate the effectiveness of FCRNet, we compared FCRNet with some similar and well-known methods, including DCAN \cite{chen2016dcan}, Discriminative Learning \cite{de2017semantic} and Harmonic Embedding \cite{kulikov2020instance}, at both pixel and object levels. All results are reported based on the same dataset.  

\section{Results}
Multiple evaluation criteria are used to benchmark the performance of FCRNet. Ensemable Dice (Dice2) \cite{vu2019methods} and Aggregated Jaccard Index (AJI) \cite{kumar2017dataset} take the split and merge errors into consideration by calculating the overlap ratio between the prediction and ground truth. F1-score \cite{caicedo2019evaluation} and Panoptic Quality (PQ) \cite{kirillov2019panoptic} are applied to inspect the segmentation at object level.

In Table \ref{table:result}, FCRNet shows superior performance on scores AJI of 0.7513 and PQ of 0.7753, and there is only a marginal difference between FCRNet and the best method from other scores. In order to set forth the efficiency of FCRNet, we also list the number of trainable parameters in different models. A compact model is preferable when there is a lack of training data. Although the Harmonic Embedding achieves the highest F1-score, it has much more parameters than that of FCRNet, increasing the demands on the dataset and training process. Furthermore, because the clustering is inherently merged into the embedding learning stage, the postprocess of FCRNet can handle about 5 images per second, which is substantially more efficient than the others. It is worth noting that the number of output channels, or colors, is essential for preventing the over-or-under segmentation. When four colors are already enough to distinguish neighboring objects, according to the four color theorem, the number of output options should not be too tight or relaxed. Otherwise it will either suppress the expressive power (like FCRNet\_3) or instigate the overfitting (like FCRNet\_10). The segmentation examples also confirm the improvements of FCRNet (Fig.\ref{fig:compare}).

\section{Conclusion}
\label{sec:typestyle}
In this work, we describe a framework FCRNet to incorporate the clustering constraint into the learning process. Explicit cluster information helps the embedding model output clustering-friendly features and guide the subsequent process.  The instance segmentation task can be formed as a one-stage optimization process, preventing the segmentation error accumulating at suboptimal modules.  Furthermore, the number of parameters is considerably reduced because there is no need for additional branches, such as clustering seeds. 

\section{Acknowledgement} This work is supported by Hong Kong Institute for Data Science.


\bibliographystyle{IEEEbib}
\bibliography{strings,refs}

\end{document}